\documentclass[11pt]{article}

\usepackage[final]{acl}
\usepackage{times}
\usepackage{latexsym}
\usepackage{xcolor}
\usepackage{graphicx}
\graphicspath{{latex/images/}}
\usepackage[T1]{fontenc}
\usepackage{todonotes}
\usepackage[utf8]{inputenc}
\usepackage{microtype}
\usepackage{inconsolata}
\usepackage{ragged2e}
\usepackage{amsmath}
\usepackage{booktabs} 
\usepackage{multirow} 
\usepackage[table]{xcolor} 
\usepackage{makecell}   
\usepackage{adjustbox}
\usepackage{xcolor}
\usepackage{booktabs}
\usepackage{enumitem}

\definecolor{strategyRed}{HTML}{B33A3A}
\definecolor{strategyBlue}{HTML}{1F5AA6}
\definecolor{strategyGreen}{HTML}{2E7D5B}

\usepackage{graphicx}

\title{\textit{Aligning with Lived Experience}: Heterogeneous Benefits of Fine Tuning in Mental Health Support Generation}

\author{Mohit Chandra$^{1}$, Nabin Kim$^{1}$, Eli Min$^{1}$, Aamogh Sawant$^{1}$, \\
  \textbf{Tanmay Sutar$^{2,}$\thanks{Work completed while at the Georgia Institute of Technology.}, Munmun De Choudhury$^{1}$} \\
  $^{1}$Georgia Institute of Technology \\
  $^{2}$Magic Hour \\
  \texttt{\{mchandra9,nkim325,elmmo,asawant43\}@gatech.edu} \\
  \texttt{tanmay.sutar@outlook.com, munmun.choudhury@cc.gatech.edu}
}

\begin{document}
\setkeys{Gin}{draft=false}

\definecolor{crimson}{rgb}{1.0, 0.1, 0.2}
\definecolor{electric blue}{rgb}{0.0, 0.5, 1.0}
\definecolor{vibrant green}{rgb}{0.0, 0.85, 0.3}
\definecolor{bright purple}{rgb}{0.7, 0.2, 1.0}
\definecolor{gold}{rgb}{1.0, 0.7, 0.0}
\definecolor{bright peach}{rgb}{1.0, 0.6, 0.45}

\maketitle
\begin{abstract}
As access to professional mental healthcare remains limited, many individuals turn to online platforms such as Reddit to seek peer support situated within human lived experience. However, a significant portion of such queries go unanswered, presenting an opportunity for using Large Language Models (LLMs) to fill this gap. While LLMs have demonstrated strong performance on clinical benchmarks, their ability to generate lived-experience informed and community-aligned peer support is underexplored. Addressing this gap, we introduce the~\textbf{CO}mmunity-centered \textbf{P}eer \textbf{E}ngaged \textbf{Support} (COPES) dataset and a three-axis evaluation framework to assess LLM alignment with community perspectives to mental health support seeking queries. Evaluating zero-shot and post-trained (SFT and DPO) models, we show that post-training on COPES significantly improves Strategy Alignment (>50\% for general-purpose models) and alignment in Emotion \& Tone. However, we also observe that such improvements are heterogeneous and alignment improvements vary significantly across subreddits and requested coping strategies. Furthermore, post-training induces distributional shifts, heavily favoring problem-focused recommendations while suppressing emotion-focused strategies. Together, this work shows that while curating community-driven data improves the alignment of LLM responses, model performance remains disparate across distinct sub-communities and specific mental health needs.
\end{abstract}

\section{Introduction}

Mental health-related conditions are increasingly prevalent in the U.S. and worldwide. It is estimated that 23.1\% of the U.S. adult population struggles with some form of mental health condition~\cite{nihMentalIllness}.~However, access to adequate care remains inaccessible, with around 48\% of U.S. adults with diagnosed conditions reporting that they were unable to obtain treatment due to affordability concerns or lack of available treatment options~\cite{samhsasurvey}.~Faced with the limitations of the healthcare system, many individuals turn to online communities to seek accessible, lived-experience-based peer support to supplement or replace professional help~\cite{Naslund2020-eo,https://doi.org/10.1111/papt.12222,doi:10.1177/2167702612463566,10.1145/2556288.2557214}.~However, past research has shown that almost 35\% of posts on health-related subreddits go unanswered~\cite{10.1145/3462204.3481748}.

Recent advancements in LLMs have positioned these systems as a tool for democratizing access to healthcare-related information and diagnosis~\cite{https://doi.org/10.1002/lrh2.70093}, with over 230 million individuals accessing ChatGPT alone for health and wellness related queries each week~\cite{openaiIntroducingChatGPTHealth}. LLMs have also demonstrated the ability to answer informational queries on health-related topics~\cite{yang-etal-2023-towards, Singhal2025}, and recent works have shown that LLM support on mental health related queries improves through post-training on base models~\cite{alghamdi2025redditessmentalhealthsocial,zhang-etal-2025-decoupledesc}.

Despite these advancements, LLMs continue to struggle in scenarios that involve knowledge about human lived experiences~\cite{schroder2025largelanguagemodelssimulate,10.1145/3715275.3732063,chandra-etal-2025-lived} and adjusting to diverse perspectives~\cite{Park2024-bb}.~This is a critical limitation when considering the usage of LLMs for providing mental health related peer support. Within this context, an effective response should be grounded in human lived experience while following community norms and expectations. Failing to align with these community-centered expectations can  cause harm to both the support-seeker and the broader community. It also remains unclear if models (zero-shot or post-trained) inadvertently provide helpful and actionable advice for certain communities within Reddit inequitably.Moreover, there is need to understand the extent to which post-training alters the types of coping strategies recommended, and whether these shifts vary based on the target community and the underlying issues discussed in the post. To fill this gap, we present the following research questions:

\vspace{0.2em}
\noindent\textbf{RQ1}:~\textit{How can we steer LLMs to generate community-centered responses for mental health queries, and how well do these models align with lived-experience perspectives?}

\vspace{0.2em}
\noindent\textbf{RQ2}: \textit{How does steering LLMs toward community-centered support affect their performance across:}
\begin{itemize}
    \vspace{-0.5em}
    \setlength{\itemsep}{0pt}
    \setlength{\parskip}{0pt}
    \item[\textbf{(a)}] \textit{posts expressing or seeking specific coping strategies?}
    \item[\textbf{(b)}] \textit{diverse mental health communities on Reddit?}
\end{itemize}

To address these research questions, we present the \textbf{CO}mmunity-centered \textbf{P}eer \textbf{E}ngaged \textbf{Support} (COPES) dataset, and an evaluation framework consisting of three axes: (1) Support Strategy Alignment, (2) Strategy Congruence, and (3) Response Emotion \& Tone for assessing LLM response alignment. Furthermore, we analyze models post-trained on COPES using Supervised Fine-Tuning (SFT) and Direct Preference Optimization (DPO) under this evaluation framework and compare performance to zero-shot baselines. We further examine performance based on the structure and content of advice provided in LLM responses and how these vary across subreddit categories.

Our findings show that post-training LLMs on the COPES dataset allows LLMs to generate recommendations grounded in lived experiences. We observed that post-training configurations (SFT and SFT+DPO) led to significant improvement in Strategy Alignment by over 50\% for general-purpose models (Qwen 3-4B Instruct, Gemma-4-E4B) and by 4.54\% for MediPhi-Instruct (SFT). However, improvements varied depending on context and models involved. Moreover, post-training led to better Emotion \& Tone Alignment for Gemma and MediPhi compared to their zero-shot baseline, while subreddits discussing topics around Psychosis \& Anxiety and Coping \& Therapy yielded better average strategy alignment across models in general. For RQ2, posts related to meaning-focused coping strategies exhibited the highest gains in Strategy Alignment (Gemma +74.27\%, Qwen +69.44\%), while meaning-focused posts showed the highest improvements in Emotion \& Tone alignment for Gemma.

Finally, post-training adjusted the coping strategies recommended by each model. Under zero-shot baseline, each model primarily used self-soothing and reframing as the uniform approach to all mental health queries. After post-training, each model developed preferred coping strategies based on the type of challenge faced. In some cases, such as posts around trauma, all three models converged on their preferred coping strategies for responses. Overall, this work presents a lived-experience centered framework for curating data and post-training models for mental health peer support task. While we observed improvements across evaluation axes, post-trained model performance remains heterogeneous, varying across both the type of support requested and the specific subreddit types.

\section{Data Collection and Curation}
\label{sec:data_collection_curation_labeling}

\begin{figure*}[!t]
    \centering
    \includegraphics[width=2\columnwidth]{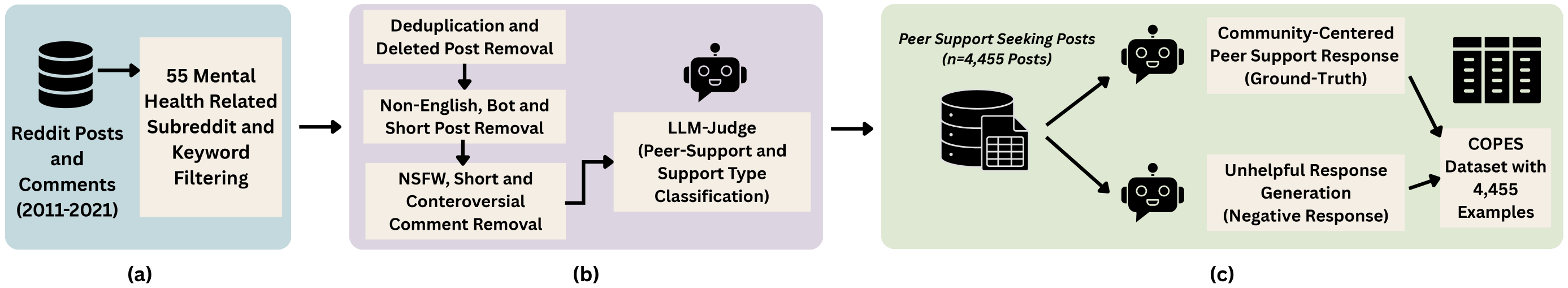}
    \caption{
Overview of the dataset construction pipeline, including Reddit thread collection, quality filtering and LLM-based peer-support/coping strategy labeling, and preferred/rejected response generation. Detailed criteria and prompts are provided in Section~\ref{sec:data_collection_curation_labeling}.
}
    \label{fig:pipeline}
\end{figure*}

In this section, we outline the dataset creation process in three stages: data collection (Section~\ref{subsec:data_collection}); pre-processing (Section~\ref{subsec:data_filtering_labeling}); and post response generation (Section~\ref{subsec:data_response_generation}).

\subsection{Reddit Thread Data Collection}
\label{subsec:data_collection}

We designed a multi-stage data collection and filtering pipeline (see Figure~\ref{fig:pipeline}) to capture authentic peer-support interactions in online mental health support communities. Using the Pushshift dataset~\cite{baumgartner2020pushshiftredditdataset}, we collected Reddit submissions and comments between 2011-2021 from mental health subreddits identified by~\citet{subreddit_lists} (see Table~\ref{tab:subreddits}). Following the taxonomy proposed by these authors, we grouped the subreddits into five community type categories: (1) Trauma \& Abuse ($C_1$), (2) Psychosis \& Anxiety ($C_2$), (3) Compulsive Disorders ($C_3$), (4) Coping \& Therapy ($C_4$), and (5) Mood Disorders ($C_5$). We additionally used a keyword-based filter to retain posts containing at least one term from a curated set of mental health and support-seeking keywords within post text or title (see Table~\ref{tab:unified_keywords}).

To improve data quality, we applied a series of filtering steps. First, we removed structurally invalid content, including posts that were deleted, removed, or duplicates. Second, we excluded non-English posts and removed non-human content by removing posts connected to known bot accounts (e.g., \textit{AutoModerator}) using regex-based patterns such as ``I’m a bot” or ``This action was performed automatically.” Third, we excluded potentially low-quality or unhelpful interactions, including NSFW content, comments shorter than five words, controversial comments (controversiality = 1), and collapsed comments. Finally, because our response-generation procedure synthesizes peer support from multiple community comments, we required posts to have at least three non-bot comments. We also retained comments with a net upvote score of $\geq 1$, treating positive community feedback as a weak endorsement signal. After filtering, 5,536 posts remained.

\subsection{Automated Data Labeling and Filtering}
\label{subsec:data_filtering_labeling}

After initial data collection and filtering (Figure~\ref{fig:pipeline}(a) and (b)), we used an LLM-as-a-judge methodology to identify posts seeking non-clinical peer support. We used GPT-4.1~\cite{gpt4-1} as the annotation model using deterministic decoding (temperature = 0.0, top-p = 1.0). We structured the automated labeling task using a two-tiered hierarchical annotation consisting of (1) a binary peer-support classification and (2) a multi-class categorization of coping strategies.

\vspace{0.1em}
\noindent\textbf{Binary Peer-Support Classification}: Within this task, we labeled a post as peer support ``yes'' if it explicitly or implicitly sought emotional encouragement, shared experiences, reassurance, or coping advice, and ``no'' if it was purely informational, technical, or non-interactive (prompt in Table~\ref{tab:peer_support_prompt}).

\vspace{0.1em}
\noindent\textbf{Coping Strategy Classification}: To characterize the different ways Reddit users express the need for peer support, we took inspiration from established psychological coping frameworks~\cite{meaning_coping_1, meaning_coping_2, problem_and_emotion, social} and categorized support types into four broad categories: (1) problem-focused, (2) emotion-focused, (3) meaning-focused, and (4) social coping. Problem-focused coping focuses on taking practical action to change external factors causing stress, while meaning-focused coping focuses on changing internal factors to reduce stress by finding personal growth in the situation. Social coping involves turning to others for support and advice, and emotion-focused coping means engaging in self-soothing behaviors to reduce distress.

For each Reddit post seeking peer support, we modeled each coping strategy category as an independent binary decision. We chose this approach over a single multi-class label to capture how multiple coping strategies often co-occur within a single post (prompt in Table~\ref{tab:coping_prompt_combined}). Additionally, to ensure robustness, we complemented automated labeling with human evaluation on a sampled subset of the data in Appendix~\ref{app:human_eval_coping}. 

After filtering out posts with missing information, we had 4,455 peer-support seeking Reddit posts with coping strategy classification labels. We further split the dataset into Train, Validation, and Test in the ratio of 70:15:15.~Table~\ref{tab:dataset_distribution} provides statistics and distribution of support strategy labels.

\begin{table}[!h]
    \centering
    \begin{adjustbox}{width=0.75\columnwidth}
    \begin{tabular}{l|ccc}
        \hline
        \textbf{Category} & \textbf{Train} & \textbf{Val} & \textbf{Test} \\
        \hline
        Total Examples    & 3118 & 669 & 668 \\
        Problem Focused   & 1151 & 224 & 233 \\
        Emotion Focused   & 2224 & 474 & 466 \\
        Meaning Focused   & 760  & 148 & 138 \\
        Social Coping    & 2115 & 451 & 472 \\
        \hline
    \end{tabular}
    \end{adjustbox}
    \caption{COPES dataset distribution of categories across data splits. Dataset was split in the ratio 70:15:15 for Train:Validation:Test. }
    \label{tab:dataset_distribution}
\end{table}

\subsection{Dataset Response Generation}
\label{subsec:data_response_generation}

For each labeled post, we created a preferred response and a rejected response, which together formed the preference pair used for post-training.

\vspace{0.1em}
\noindent\textbf{Preferred Response Generation (Ground-Truth)}: We constructed the preferred response by summarizing the peer support provided in the Reddit comments for each post. We used GPT-4.1 as an LLM-based summarizer to combine the associated comments for each post into a single coherent response. The prompt also asked the model to remove redundant content, avoid generic openings, and maintain a natural peer-support tone (see Table~\ref{tab:positive_answer_prompt}).

\vspace{0.1em}
\noindent\textbf{Rejected Response Generation}: To construct preference pairs for Direct Preference Optimization (DPO)~\cite{DPO}, we also created a non-preferred response (rejected response) for every post in the dataset. Our initial experiments indicated that summarizing downvoted comments for non-preferred responses was unreliable, as these often reflected engagement rather than relevance. Hence, we generated rejected response variants using GPT-4.1 to synthesize suboptimal user responses from the original Reddit threads (prompt in Table~\ref{tab:negative_prompt}). To encourage semantic and stylistic diversity among rejected responses, we defined a theoretically situated taxonomy of 11 negative support strategies (see Table~\ref{tab:negative_taxonomy}), including emotional invalidation, misguided positivity, and social undermining, grounded in prior work on harmful social support and communication~\cite{zielinski2018_emo_invalid, vinokur1993_soc_undermine, revenson1991_ctrl_support, almajed2026incongruentpositivitymiscalibratedpositivity, feng2016_unsol_advice, vangelisti1990_emotional_displacement, rozin1999_moralize, pargament2003_religious_coping}.

\section{Model Selection and Evaluation Axes}
\label{sec:model_selection_evaluation}

In this section, we present our evaluation and post-training methodology for LLMs to provide peer-like, long-form support and advice. We used two general-purpose models: (1) Qwen 3-4B Instruct~\cite{yang2025qwen3technicalreport}, (2) Google Gemma-4-E4B~\cite{deepmindGemma4}, and one model specific to the medical domain, (3) Microsoft MediPhi-Instruct (3.8B)~\cite{corbeil-etal-2025-modular}. We selected these models to compare general-purpose and medical-domain instruction-tuned models under a comparable small-model setting. Additionally, their accessible weights allowed us to apply the same SFT and DPO pipeline while avoiding dependence on proprietary model APIs. For the evaluation criteria we used three metrics: (1) Support Strategy Alignment, (2) Support Strategy Congruence, and (3) Emotion \& Tone Alignment. 

\vspace{0.1em}
\noindent\textbf{Support Strategy Alignment}: Reddit users frequently seek peer support grounded in lived human experience~\cite{bbcRedditsHuman}. We therefore focus on whether generated responses preserve coping advice reflected in ground-truth responses. We operationalized support strategy alignment as a proxy measure of whether LLM-generated responses preserve coping recommendations reflected in ground-truth preferred responses, not as a direct measure of overall peer-support quality. Given the long-form nature of both ground-truth and LLM-generated responses (prompt in Table~\ref{tab:answer_generation_prompt}), assessing their alignment required us to disentangle the semantic content from the stylistic features. Hence, we extracted the recommendations present in the ground truth and the LLM-generated responses. We treated each extracted recommendation as an atomic support strategy. Then, we used vector embeddings to match the LLM-generated support strategies with the ground truth strategies. We describe both in detail below. 

\begin{table*}[!t]
\centering
\begin{adjustbox}{width=\linewidth}
\renewcommand{\arraystretch}{1.3} 
\begin{tabular}{c ccc ccc ccc}
\toprule
& \multicolumn{3}{c}{\textbf{Strategy Alignment}} & \multicolumn{3}{c}{\textbf{Strategy Congruence}} & \multicolumn{3}{c}{\textbf{Emotion \& Tone Alignment}} \\
\cmidrule(lr){2-4} \cmidrule(lr){5-7} \cmidrule(lr){8-10}
\textbf{Model} & Zero-Shot & SFT & SFT+DPO & Zero-Shot & SFT & SFT+DPO & Zero-Shot & SFT & SFT+DPO \\
\midrule
\rowcolor{gray!10} \makecell[l]{Qwen 3-4B \\ Instruct} & \makecell{0.30 \\  $\pm$ 0.284} & \makecell{0.47 \\  $\pm$ 0.320 \\ \small (+56.52\%)***} & \makecell{0.48 \\  $\pm$ 0.318 \\ \small (+60.32\%)***} & \makecell{0.64 \\ $\pm$ 0.050} & \makecell{0.65 \\ $\pm$ 0.059 \\ \small (+1.71\%)***} & \makecell{0.65 \\ $\pm$ 0.058 \\ \small (+2.06\%)***} & \makecell{2.73 \\ $\pm$ 4.296} & \makecell{2.80 \\ $\pm$ 4.071 \\ \small (-2.33\%)} & \makecell{2.78 \\ $\pm$ 4.147 \\ \small (-1.63\%)} \\

\makecell[l]{Gemma-4-E4B} & \makecell{0.32 \\  $\pm$ 0.284} & \makecell{0.50 \\  $\pm$ 0.318 \\ \small (+55.06\%)***} & \makecell{0.50 \\  $\pm$ 0.317 \\ \small (+54.59\%)***} & \makecell{0.62 \\ $\pm$ 0.055} & \makecell{0.65 \\ $\pm$ 0.056 \\ \small (+3.63\%)***} & \makecell{0.65 \\ $\pm$ 0.060 \\ \small (+3.40\%)***} & \makecell{3.23 \\ $\pm$ 4.640} & \makecell{2.79 \\ $\pm$ 4.161 \\ \small (+13.51\%)***} & \makecell{2.94 \\ $\pm$ 4.729 \\ \small (+9.04\%)***} \\

\rowcolor{gray!10} \makecell[l]{MediPhi-Instruct } & \makecell{0.46 \\  $\pm$ 0.345} & \makecell{0.48 \\  $\pm$ 0.320 \\ \small (+4.54\%)} & \makecell{0.47 \\  $\pm$ 0.326 \\ \small (+3.03\%)} & \makecell{0.65 \\ $\pm$ 0.058} & \makecell{0.65 \\ $\pm$ 0.061 \\ \small (-0.92\%)***} & \makecell{0.65 \\ $\pm$ 0.057 \\ \small (-0.93\%)**} & \makecell{2.76 \\ $\pm$ 4.951} & \makecell{2.65 \\ $\pm$ 4.205 \\ \small (+3.78\%)} & \makecell{2.73 \\ $\pm$ 4.142 \\ \small (+1.06\%)} \\
\bottomrule
\end{tabular}
\end{adjustbox}
\caption{Evaluation of response alignment for LLMs across different training configurations. The table compares Strategy Alignment, Strategy Congruence, and Emotion \& Tone Alignment for baseline (Zero-Shot), SFT, and SFT+DPO configurations. Values represent mean scores $\pm$ standard deviation. Percentage values in parentheses indicate relative improvement over the Zero-Shot baseline, with * denoting statistical significance. For Strategy Alignment and Emotion \& Tone alignment we used Wilcoxon Signed-Rank Test for checking significance of the results. For Strategy Congruence we used Mann-Whitney U Test. }
\label{tab:rq_1_metric_eval_table}
\end{table*}

Drawing on the definition provided by~\citet{min2023factscore}, we framed atomic support strategies as short, self-contained recommendations that express one coping suggestion at a time. We used the Qwen3-30B-A3B-Instruct-2507 model~\cite{yang2025qwen3technicalreport} and the prompting methodology described by~\citet{chandra-etal-2025-lived} for the strategy extraction process (prompt in Table~\ref{tab:atomic_extraction}). To reduce redundancy among the extracted atomic strategies, we combined strategies based on the actors and actions involved (prompt in Table~\ref{tab:atomic_combination}). We grouped the combined strategies into the four coping categories introduced in Section~\ref{subsec:data_filtering_labeling}: Emotion-Focused Coping, Problem-Focused Coping, Meaning-Focused Coping, and Social Coping~\cite{folkman-moskowitz-2004-coping, algorani-gupta-2023-coping}. We validated the extraction and combination steps using human evaluation over a random sample of 30 responses, which included 187 extracted strategies and 100 combined groups. Two annotators independently labeled these results and achieved an inter-rater agreement score of 0.90 using Cohen's Kappa. Examples of extracted strategies and their combined groups are available in Table~\ref{tab:atomic_comparison} in the Appendix.

To evaluate strategy alignment between the LLM-generated and ground truth responses, we constructed a vector database containing embeddings of Reddit post text, title, and extracted support strategies from the ground truth answer using Qwen3-Embedding-4B model. Each entry also contained information about the respective subreddit category. For each atomic support strategy ($s_{k}$) extracted from an LLM response ($L_t$) corresponding to a post text within the test set ($P_t$), we first used the thematic similarity within subreddits in the same category ($C_i$) by isolating all posts in the same sub-category as $P_t$. We then compute the cosine similarity between the embedding of the test set post ($P_t$) and the embeddings of all posts in this filtered subset, retaining candidates that meet a specific similarity score threshold $\geq0.8$ and capping the results at the top 10 most similar posts to ensure maximum relevance. Finally, for each extracted strategy $s_{k}$ from the LLM response ($L_t$), we compare its embedding with the strategy embeddings of the retrieved candidate posts' Ground-Truth answers. A strategy is considered ``aligned'' if it achieves the required cosine similarity threshold $\geq0.8$ with one or more strategies in the candidate set.

\vspace{0.1em}
\noindent\textbf{Support Strategy Congruence}: In addition to support strategy alignment, we  analyzed how semantically coherent the extracted strategies are within each LLM-generated response. For a given LLM response $L_t$ containing a set of strategy embeddings $\{s_1, s_2, \dots, s_n\}$, we define \textbf{congruence} as the mean pairwise cosine similarity among the extracted strategy embeddings. Higher congruence indicates that a response provides a more semantically coherent set of recommendations. However, a marginal reduction in congruence is not necessarily detrimental for the task of providing peer support, where diverse perspectives are often valued.

\vspace{0.1em}
\noindent\textbf{Emotion and Tone Alignment}: Previous work has highlighted the importance of emotional alignment between the support provider and support seeker to drive meaningful change~\cite{Atzil_Slonim_2018}. Inspired by prior work, we used Empath~\cite{10.1145/2858036.2858535} to measure emotion and tone alignment between LLM-generated responses and ground truth along 8 relevant emotional and tonal categories~\cite{chandra-etal-2025-lived}. We analyzed the distribution of these categories and measured alignment using Kullback-Leibler (KL) divergence.

\section{RQ1: Peer-Support Response Generation}

Table~\ref{tab:rq_1_metric_eval_table} reports Strategy Alignment, Strategy Congruence, and Emotion \& Tone Alignment across zero-shot, SFT, and SFT+DPO configurations for all three models.

\subsection{Evaluation of Zero-shot and Post-Trained Models on COPES}

\noindent\textbf{Strategy Alignment}: As shown in Table~\ref{tab:rq_1_metric_eval_table}, strategy alignment proved to be a challenging objective across all models (zero-shot or post-trained). The maximum strategy alignment was achieved for Gemma-4-E4B under the SFT method ($50\%$). However, post-training using SFT and SFT+DPO resulted in improvement over the zero-shot baseline across all models. Specifically, we observed the largest gain in alignment for Qwen 3-4B with SFT (and SFT+DPO) increasing alignment by $56.52\%$ ($60.32\%$). In contrast, MediPhi-Instruct showed lower improvement of $4.54\%$ for SFT and $3.03\%$ for SFT+DPO. These results reinforce the challenges of Personalized Alignment \cite{guan-etal-2025-survey}, especially for tasks involving nuanced aspects of human social and emotional scenarios such as seeking peer support.

Although SFT and SFT+DPO produced large relative gains, the absolute strategy alignment scores remained modest, suggesting that post-training only partially captures the community context underlying Reddit peer support. This may reflect the heterogeneity of peer-support interactions, where responses can involve advice, validation, disclosure, and shared lived experience rather than a single discrete coping strategy. This limitation echoes broader work on prediction in complex social systems, where performance can be constrained by the inherent variability of social behavior as well as by data or model quality \citep{martin2016exploring}.

\vspace{0.2em}
\noindent\textbf{Strategy Congruence}: We observed modest improvements for the general-purpose models, with Gemma-4-E4B achieving improvement of $3.63\%$ and $3.40\%$ using SFT and SFT+DPO respectively. In contrast, MediPhi-Instruct showed a decrease of $0.93\%$ (SFT+DPO) and $0.92\%$ (SFT) compared to the baseline zero-shot setting. A plausible explanation for this observation could be the ceiling effect related to thematic consistency of recommendations. MediPhi-Instruct training on medical corpora likely established a robust zero-shot baseline for peer-support tasks. Hence, this high initial baseline left limited room for substantial post-training gain in strategy congruence. Another plausible explanation is that diverse peer-support perspectives may be inconsistent with standard clinical guidelines found in biomedical literature used for training medical LLMs.

\begin{figure*}[!t]
    \centering
    \includegraphics[width=2\columnwidth]{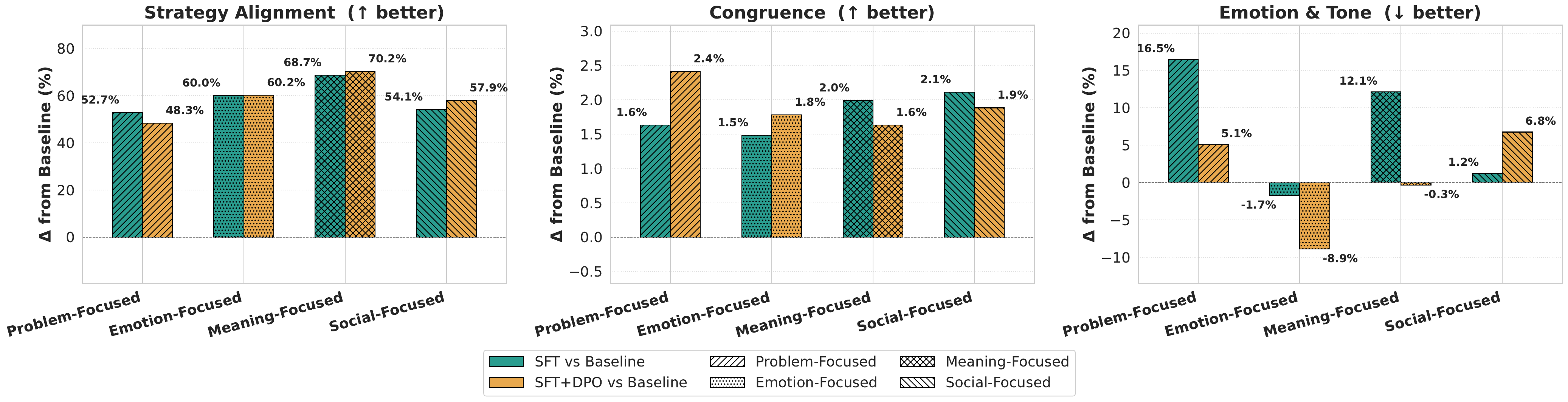}
    \caption{Relative changes in Strategy Alignment, Strategy Congruence, and Emotion \& Tone metrics for Qwen-3 4B following SFT and SFT+DPO, compared to zero-shot baselines across Reddit post coping strategy labels. Results for Gemma, Mediphi and Qwen are presented in Appendix Figure~\ref{fig:appendix_rq2_coping_difference_full}.}
    \label{fig:rq2_coping_strategy_performance_qwen}
\end{figure*}

\vspace{0.2em}
\noindent\textbf{Emotion and Tone Alignment}: Post-training generally reduced the KL divergence of the category distribution for LLM responses to the preferred responses distribution (with exception of Qwen 3-4B Instruct). While we observed no significant difference in the distribution of categories within LLM responses with the distribution within ground truth under any setting, we observed that Gemma model produced notable more aligned responses with the ground truth when compared to the zero shot baseline (+13.51\% for SFT and 9.04\% for SFT+DPO).

\subsection{Performance Disparity for Different Types of Subreddits}

\begin{table}[htbp]
    \centering
    \begin{adjustbox}{width=\columnwidth}
    \setlength{\tabcolsep}{2.5pt}

    \begin{tabular}{lcccccc}
        \toprule
        & \multicolumn{2}{c}{\makecell[c]{\textbf{Gemma-4}\\\textbf{E4B}}} & \multicolumn{2}{c}{\makecell[c]{\textbf{Qwen 3-4B}\\\textbf{Instruct}}} & \multicolumn{2}{c}{\makecell[c]{\textbf{MediPhi-Instruct}\\\textbf{Instruct}}} \\
        \cmidrule(lr){2-3} \cmidrule(lr){4-5} \cmidrule(lr){6-7}
        \textbf{Cat.} & \makecell[c]{\textbf{SFT}} & \makecell[c]{\textbf{SFT+}\\\textbf{DPO}} & \makecell[c]{\textbf{SFT}} & \makecell[c]{\textbf{SFT+}\\\textbf{DPO}} & \makecell[c]{\textbf{SFT}} & \makecell[c]{\textbf{SFT+}\\\textbf{DPO}} \\
        \midrule
        C1 & \cellcolor{teal!54}89.98\% & \cellcolor{teal!62}102.84\% & \cellcolor{teal!10}16.03\%  & \cellcolor{teal!12}20.30\%  & \cellcolor{teal!17}28.44\%  & \cellcolor{teal!12}19.53\%   \\
        C2 & \cellcolor{teal!41}68.97\%  & \cellcolor{teal!46}77.34\%  & \cellcolor{teal!34}56.64\%  & \cellcolor{teal!38}63.75\%  & \cellcolor{teal!11}18.89\%  & \cellcolor{teal!12}20.47\%  \\
        C3 & \cellcolor{teal!12}20.48\%  & \cellcolor{teal!7}11.45\%  & \cellcolor{teal!65}340.00\%  & \cellcolor{teal!65}397.78\%  & \cellcolor{orange!5}-0.15\%  & \cellcolor{orange!26}-17.02\%   \\
        C4 & \cellcolor{teal!33}54.29\%  & \cellcolor{teal!26}43.29\%  & \cellcolor{teal!51}84.65\%  & \cellcolor{teal!60}99.34\%  & \cellcolor{teal!6}9.68\%  & \cellcolor{teal!6}10.35\%  \\
        C5 & \cellcolor{teal!28}46.56\%  & \cellcolor{teal!26}43.82\%  & \cellcolor{teal!31}52.41\%  & \cellcolor{teal!31}51.48\%  & \cellcolor{orange!7}-4.93\%   & \cellcolor{orange!11}-7.53\%   \\
        \bottomrule
    \end{tabular}
    \end{adjustbox}
    \caption{Relative improvement (\%) in strategy alignment for post-trained model variants compared to each model's zero-shot baseline across subreddit categories ($C_1 \dots C_5$; Ref:~\ref{subsec:data_collection}). Colors indicate direction of change (teal for positive, orange for negative).}
    \label{tab:rq1_subreddit_improvement_analysis}
\end{table}

We further evaluated whether post-training led to varied improvements across the five subreddit communities ($C_1 \dots C_5$). Table~\ref{tab:rq1_subreddit_improvement_analysis} shows that post-training generally improved the strategy alignment scores across all models and categories except in the case of MediPhi-Instruct for $C_5$ and $C_3$ subreddit categories. On average, the relative improvement after post-training (compared to baseline) was most pronounced for subreddits related to $C_2$ (Psychosis \& Anxiety) and $C_4$ (Coping \& Therapy) communities (excluding Qwen 3 performance for $C_3$). On the other hand, when treating the unusually large Qwen gains for C3 as an outlier, we observed the least average performance boost across the model families for $C_5$ (Mood Disorders). Within $C_1$ (Trauma \& Abuse), smaller and more model-dependent gains may reflect the fact that trauma-related support often foregrounds safety, validation, disclosure, and lived experience, which are less likely to appear as explicit, extractable coping strategies.

\section{RQ2: Impact of Post-Training on Response Generation for Varied Coping Strategies}

For the second research question, we analyzed how post-training effects vary across coping strategies and subreddit communities through three complementary analyses. First, we used the coping strategy labels associated with each test-set post (Section~\ref{subsec:data_filtering_labeling}) to analyze the heterogeneity in model performance across the four coping strategies under SFT and SFT+DPO relative to the zero-shot baseline. Second, we examined how post-training changed the distribution of generated coping strategy types within the LLM response. Additionally, we also measured broader topical alignment between LLM generated responses and the ground-truth responses using a Hierarchical Dirichlet Process (HDP) model. Specifically, we adopted the $sim_{\text{HDP}}$ metric from~\citet{10.1145/3589334.3645643}. For each model and training configuration, we fitted an HDP model to the generated responses and computed the cosine similarity between each output and its reference in the same topic space.

\subsection{Performance Heterogeneity for Varied Coping Strategy Types}

Across all models and coping strategy categories, SFT and SFT+DPO based post-training improved Strategy Alignment compared to the zero-shot baseline (except for Mediphi for emotion focused coping). For SFT+DPO setting, general-purpose models achieved substantially larger relative gains than MediPhi-Instruct across categories, with Gemma-4-E4B and Qwen 3-4B (Figure~\ref{fig:rq2_coping_strategy_performance_qwen}) improving by +60.35\% and +59.13\% on average across all coping categories, respectively, compared to +7.02\% for MediPhi-Instruct. However, the gain in performance was unevenly distributed across the coping strategy categories. For both Gemma and Qwen, posts annotated with meaning-focused coping showed the highest average gains across post-training methods (+74.27\% and +69.44\%, respectively). This pattern may indicate that post-training better captures meaning-focused support.

While we observed similar gain in Strategy Congruence scores for Gemma and Qwen models across coping strategy categories, MediPhi-Instruct showed a marginal decrease ($<$2.5\%) in most categories, except for problem-focused coping under the SFT setting. For Emotion \& Tone alignment, we observed that SFT configuration led to the largest improvements for meaning-focused coping posts for Gemma (+19.33\%). On the other hand, MediPhi-Instruct exhibited the greatest increased in emotion and tone alignment for problem-focused coping posts (+12.68\% under SFT and +17.07\% under SFT+DPO).

\subsection{Change in Coping Strategy Type Distribution}

For the zero-shot setting, general-purpose model responses focused on suggesting self-soothing and reframing coping strategies. Emotion-focused approaches appeared in an average of 35.3\% of responses for Qwen and up to 55.1\% for Gemma. In contrast, MediPhi generally maintained a uniform approach to each subreddit category, suggesting that almost every issue was receiving at least one recommendation from each coping strategy. 

\begin{figure*}[!t]
    \centering
    \includegraphics[width=1.8\columnwidth]{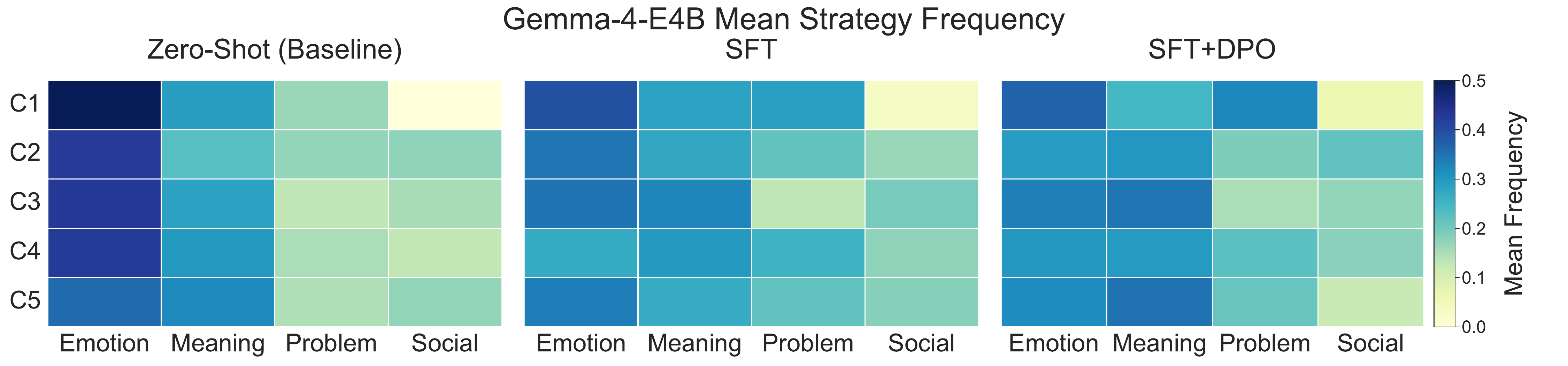}
    \caption{Distribution of generated coping strategy types across subreddit community categories for Gemma-4 E4B. Post-training changes the type of advice models produce, with general-purpose models shifting toward more problem-focused recommendations. Results for Gemma, Mediphi and Qwen are presented in Appendix Figure~\ref{fig:appendix_strategy_frequency_full}.}
    \label{fig:strategy_frequency_single}
\end{figure*}

After post-training, each model developed a more contextual approach to the issues presented within the Reddit posts (see Figure~\ref{fig:appendix_strategy_frequency_full}). For Trauma \& Abuse related posts ($C_1$), all three models converged on emotion-focused coping as the most frequently used strategy, shifting by +2.8\% for Qwen, +15.1\% for MediPhi, and -20.8\% for Gemma under SFT compared to their zero-shot baselines. This was followed by problem-focused coping as the second most frequently recommended strategy. However, this alignment on coping strategies did not always happen; models contained some discrepancies in their preferred strategies. Noteworthy examples were in social coping and problem-focused coping, which were emphasized in some cases and used infrequently in others. Social coping was used infrequently for $C_1$ category posts, with the largest drop being MediPhi-Instruct (-15.4\% SFT+DPO), but it was used more frequently for Psychosis \& Anxiety ($C_2$) category posts at 24.5\% for MediPhi (w.r.t. the zero-shot setting). Gemma generally did not use social coping approaches for $C_5$ (Mood Disorders) even though other models do not make this strong distinction (see Figure~\ref{fig:strategy_frequency_single}).

\subsection{Changes in Response Topical Theme}

As shown in Figure~\ref{appendix_fig:topical_theme}. After SFT, models showed mixed changes in topical alignment compared to their zero-shot baselines. Gemma-4 E4B mostly preserved topical alignment, with changes ranging from $-3.4\%$ for $C_1$ to $+9.6\%$ for $C_2$. MediPhi-Instruct showed a similar pattern, with changes ranging from $-4.8\%$ for $C_5$ to $+2\%$ for $C_2$. Both models in general preserved category-specific content while integrating peer-support language, such as validation, reassurance, and emotional grounding. In contrast, Qwen-3 4B's topical alignment declined across all five categories, with drops from $-2.4\%$ for $C_3$ to $-13.3\%$ for $C_2$. This observed shift was due to responses shifting toward broader supportive themes at the expense of reference-specific content, especially around anxiety and psychosis-related symptoms in $C_2$.

Under SFT+DPO, all three models showed lower topical alignment than their zero-shot baselines. Gemma-4B declined across every category, with the largest
drop in $C_1$ ($-32.5\%$). Similarly, Qwen-3 4B exhibited reduction in alignment, ranging from $-7.2\%$ for $C_4$ to $-16.2\%$ for $C_3$. Compared to the baseline, SFT+DPO introduced a stronger and more consistent topic shift across models. Similar to SFT, responses focused on broader peer support related topics. Overall, this analysis shows that this shift induced due to post-training is highly dependent on the target subreddit category.

\section{Related Work}

\paragraph{NLP for Mental Health Related Tasks}

Use of language models for mental health-related tasks has recently gained traction, from identifying empathic conversations~\cite{sharma-etal-2020-computational}, to generating emotional support dialogues with predefined support strategies~\cite{liu-etal-2021-towards}, or responses based on community response signals and support seeker's response~\cite{alghamdi2025redditessmentalhealthsocial}. With the rise of LLMs, benchmarking and evaluation have become one of the most-studied aspects. Previous studies have assessed LLMs for mental health counseling compentancy~\cite{nguyen-etal-2025-large}, multi-turn sensemaking~\cite{chandra2025reasoningneedexaminingllms}, adherence to motivational interviewing theory~\cite{gabriel-etal-2024-ai}, medical examination questions~\cite{Singhal2025}, and alignment with experts on psychiatric medication ADRs~\cite{chandra-etal-2025-lived}. To improve LLM performance, researchers have also used few-shot prompting for mental health condition detection ~\cite{yang-etal-2023-towards}, RAG for mental health text analysis~\cite{kermani-etal-2025-systematic}, and preference optimization for emotional support generation~\cite{zhang-etal-2025-decoupledesc}. However, past works have understudied model's ability to preserve community-specific coping strategies~\cite{guan-etal-2025-survey}. Our work addresses this gap.

\paragraph{Social media based peer support and coping}

Online social media platforms especially Reddit serve as an important avenue for individuals to talk about mental health because it supports pseudonymous self-disclosure, access to community, and large-scale mental health discourse~\cite{DeChoudhury_De_2014,cohan2018smhd,morini2025participant}. Past works have examined the relationship between linguistic accommodation and support received~\cite{subreddit_lists}, link between receiving early social support and providing support to others~\cite{10.1145/3411763.3451644} impact of language of received support on suicidal ideation risk~\cite{dechoudhury2017language}, and variation in informational and emotional support provided by users across community structures \citep{support_keywords}. Other works have also focused on characterizing effective support includes complex language factors such as diversity, adaptability, and style~\cite{Saha_Sharma_2020}. Our work builds on this literature by using Reddit peer interactions as a community-grounded reference for evaluating whether LLM-generated responses align with human lived-experiences.

\section{Conclusion}

In this work, we introduced COPES, a community-centered dataset for mental health-related peer-support tasks, and a multi-axis offline evaluation framework to examine LLM-generated peer-support response alignment with community grounded responses. Using COPES and our evaluation framework, we evaluated two general-purpose and one medical LLM across zero-shot, SFT, and SFT+DPO settings. Our findings show that post-training on COPES resulted in a significant increase in support strategy alignment compared to the baseline zero-shot version ($>50\%$) for general-purpose models. Additionally, post-training also shifted model behavior from recommending uniform self-soothing and reframing coping strategies to context-specific coping strategies. However, the benefits from post-training remain heterogeneous, varying across model architectures, the specific community context, and the nature of the support requested.

\section{Limitations}

While our work presents a novel dataset, a methodology for evaluating community-centeredness of LLM responses and novel insights on the impact of post-training, it is important to acknowledge the limitations of this work. First, the COPES dataset focuses on non-clinical mental health support-seeking behavior which does not capture the full range of clinical scenarios. To construct COPES at scale, we used LLM-as-a-judge methodology to label posts. While we conducted human evaluation to validate LLM labels, reliance on automated labeling remains a limitation. Our findings should be interpreted within the scope of the dataset construction procedure. COPES is built from answered Reddit threads collected between 2011 and 2021 from a fixed set of mental health subreddits. Because our method requires multiple positively received comments, COPES reflects well-engaged, community-endorsed peer-support exchanges rather than unanswered-post deployment. As a result, our findings may not fully generalize to unanswered posts, post-2021 community norms, or unseen mental health communities, and may over-represent threads that were more visible or socially supported.

Furthermore, we used LLMs to generate the negative responses due to the infeasibility of manually writing long-form responses for over 4,400 Reddit posts. Second, we also acknowledge the limitations in our evaluation methodology. To facilitate evaluation at scale, we measured strategy alignment using an embedding similarity approach and kept a strict threshold to reduce false positives. However, this conservative approach may sometimes miss forms of support that primarily rely on narrative disclosures or emotional validation.

We also acknowledge the limitation of evaluating only small open-weight models. This constraint was dictated by computational and budget constraints, as well as our requirement for direct access to model weights for post-training rather than reliance on proprietary model APIs. Finally, we acknowledge that our study is limited to two post-training methods (SFT and DPO). Using reward-based preference tuning methods required defining custom reward signals which fell outside the scope of this paper. Despite these limitations, our work provides a community-centered dataset, an evaluation framework for studying support-strategy alignment in LLM-generated peer support responses, and a reproducible post-training pipeline for future research.

\section{Ethical considerations}

For the creation of the COPES dataset, we used publicly available archival Reddit data from the Pushshift dataset~\cite{baumgartner2020pushshiftredditdataset}. Because our work involved retrospective analysis of public posts without direct interaction with Reddit users, the Institutional Review Board (IRB) classified it as non-human subjects research and exempted it from IRB approval. The datset is intended to be used for research purposes. We also checked the dataset for personally identifiable information using the Microsoft Presidio library, focusing on entities detected with high confidence (threshold $> 0.7$). This analysis did not identify sensitive personal identifiers, detected entities were limited to generic email addresses and helpline phone numbers. To reduce privacy risks, we will release COPES using Reddit post IDs rather than raw Reddit text. Finally, all human evaluations used to validate the LLM-generated labels and the broader evaluation pipeline were conducted by the authors, and no external participants were recruited.

COPES should be treated as a research benchmark rather than evidence of readiness for real-world mental health support deployment. Better performance on COPES should not be interpreted as evidence of a model's safety or efficacy in providing peer support. LLM-generated responses may still provide inappropriate advice, miss signs of crisis, overgeneralize from community patterns, or fail to respond to individual needs. Deploying LLMs for real-world mental health applications would require additional safeguards, crisis escalation protocols, human oversight, and evaluation with affected communities.

Finally, post-training models to produce lived-experience-informed responses introduces an important trade-off. Online communities can provide valuable peer support grounded in lived experience, but they may also propagate stigma, harmful advice, or norms that are not clinically appropriate. By synthesizing preferred responses from Reddit comments, we aimed to study coping-strategy alignment rather than evaluate the clinical efficacy of community-generated advice. A remaining challenge is how to balance community-grounded alignment with safety, clinical appropriateness, and the needs of affected users.

\section{Acknowledgments}

Chandra and De Choudhury were partly supported through awards from Microsoft and Google. Min was supported through an NSF Graduate Fellowship.

\bibliography{custom}

\clearpage
\appendix

\begin{table*}[!t]
\centering
\small
\setlength{\tabcolsep}{4pt}
\begin{tabularx}{\textwidth}{X}
\toprule
\textbf{Subreddits} \\
\midrule

r/abuse, r/adultsurvivors, r/aftersilence, r/Anger, r/bullying, r/CPTSD, r/domesticviolence, r/emotionalabuse, r/ptsd, r/PTSDCombat, r/rapecounseling, r/StopSelfHarm, r/survivorsofabuse, r/traumatoolbox, \\

r/Agoraphobia, r/Anxiety, r/BipolarReddit, r/BipolarSOs, r/BPD, r/dpdr, r/psychoticreddit, r/MaladaptiveDreaming, r/Psychosis, r/PanicParty, r/schizophrenia, r/socialanxiety, \\

r/compulsions, r/CompulsiveSkinPicking, r/OCD, r/Trichsters, \\

r/CupsOfTea, r/BackOnYourFeet, r/Existential\_crisis, r/getting\_over\_it, r/GriefSupport, r/helpmecope, r/hardshipmates, r/HereToHelp, r/iflostbetter, r/LostALovedOne, r/offmychest, r/MMFB, r/Miscarriage, r/reasonstolive, r/SuicideBereavement, r/therapy, \\

r/depression, r/depressed, r/ForeverAlone, r/GFD, r/lonely, r/mentalhealth, r/Radical\_Mental\_Health, r/SuicideWatch

\\
\bottomrule
\end{tabularx}
\caption{List of 55 mental health–related subreddits used in this study.}
\label{tab:subreddits}
\end{table*}

\begin{table*}[!t]
\centering
\small
\setlength{\tabcolsep}{4pt}
\begin{tabularx}{\textwidth}{p{0.20\textwidth} X}
\toprule
\textbf{Keyword Type} & \textbf{Keyword List} \\
\midrule
Mental Disorder Terms &
Anxiety disorder, Generalised anxiety disorder, Panic disorder, Social anxiety disorder, Separation anxiety disorder, Depression, Depressive episode, Bipolar disorder, Manic episode, Post-Traumatic Stress Disorder, PTSD, Schizophrenia, Delusions, Hallucinations, Disorganised thinking, Disorganised behaviour, Eating disorder, Anorexia nervosa, Bulimia nervosa, Disruptive behaviour disorder, Dissocial disorder, Conduct disorder, Oppositional defiant disorder, ODD, Neurodevelopmental disorder, Autism spectrum disorder, ASD, Attention-deficit/hyperactivity disorder, ADHD, Disorders of intellectual development, Intellectual disability \\
\midrule
Topic Keywords &
advice, today, need, therapist, end, never, daydream, anything, birthday, read, ocd, hate, job, question, deal, think, new, looking, interview, diagnosed, feeling, normal, tell, therapy, guy, head, thought, suicidal, better, anxiety, social, intrusive, life, symptom, tired, relationship, worst, physical, friend, scared, depressed, kill, best, love, disorder, losing, done, lonely, one, talk, take, could, wish, person, go, care, depression, bad, attack, panic, this, heart, due, weird, cause, anybody, call, work, anxious, getting, alone, wrong, school, family, back, always, mental, health, stop, fuck, even, illness, sure, compulsion, much, time, day, year, first, every, daydreaming, right, sleep, week \\
\bottomrule
\end{tabularx}

\caption{Unified keyword set used for filtering and analysis.}
\caption*{\footnotesize
Keyword filtering is widely used in mental health dataset construction to improve precision \cite{cohan2018smhd}. 
Our keyword list combines (1) mental disorder terminology derived from public health definitions \cite{who2025mentaldisorders} 
and (2) support-seeking expressions observed in online discourse \cite{support_keywords}. 
We further expand the vocabulary with lexical variations (e.g., \textit{PTSD}, \textit{anxiety}, \textit{depress}).
}
\label{tab:unified_keywords}
\end{table*}

\begin{table*}[p]
\centering
\small
\renewcommand{\arraystretch}{0.88}
\setlength{\tabcolsep}{3pt}
\begin{tabularx}{\textwidth}{l|X}
\toprule
\textbf{Type} & \textbf{Prompt} \\
\midrule

\textbf{System Prompt} & 
You are a clinical psychologist specializing in digital mental health and social support systems. Your task is to identify whether a Reddit post expresses a request for non-medical peer support.\par

Definition: Within non-medical peer support, people who share common experiences or face similar challenges come together as equals to give and receive help based on the knowledge that comes through shared experience. It focuses on validation, normalization, and shared coping rather than technical solutions or clinical diagnosis.\par

You will be given a Reddit post text (REDDIT\_POST\_TEXT) and Reddit post title (REDDIT\_POST\_TITLE) and your task is to determine whether the post is requesting non-medical, emotional peer support. Some examples are:\par

\begin{itemize}
\item Seeking emotional encouragement or reassurance.
\item Asking for shared experiences to overcome a problem.
\item Asking for connecting to people, resources, needed services, and community.
\item Asking for steps/help for overcoming difficult situations.
\item Turning to the community for finding personal growth, purpose, or a new perspective in the situation.
\end{itemize}

Do NOT classify as peer support if the post is:\par

\begin{itemize}
\item A technical or troubleshooting question
\item A medical or disease related query
\item A venting or reflection without inviting interaction/reflection
\item A game strategy or product-related coordination
\end{itemize}

\textbf{Examples:}\par

Example 1:\par
Post: ``I've been feeling numb ever since my mom passed. How do you even begin to move on?''\par
Label: YES\par
Reasoning: The user asks for advice on how to cope with grief---clearly seeking emotional guidance.\par

Example 2:\par
Post: ``Can you turn autosaves off? I accidentally ruined my build.''\par
Label: NO\par
Reasoning: This is a technical question about game mechanics, not a request for emotional support.\par

Example 3:\par
Post: ``Does anyone else feel like they're just pretending to be okay all the time?''\par
Label: YES\par
Reasoning: The user is asking for shared experiences to feel less alone, satisfying the peer support criteria.\par

Example 4:\par
Post: ``Just hit 10k steps today. Feeling good!''\par
Label: NO\par
Reasoning: This is a self-reflective statement celebrating an achievement, not seeking support.\par

Example 5:\par
Post: ``I always binge uncontrollably during my period. I beat myself up and get depressed. Anyone else? What helps you stop?''\par
Label: YES\par
Reasoning: The post contains both a request for shared experiences and a question about coping strategies.\par

Example 6:\par
Post: ``I'm looking for anyone available to help sweep through the rest of my storm shield defenses.''\par
Label: NO\par
Reasoning: This is a game coordination request, not an emotional or social support need.\par

\textbf{Instructions:}\par

You must respond strictly using this JSON format:\par

\texttt{\{}\par
\texttt{"reasoning": "your reasoning (1--2 sentences)",}\par
\texttt{"answer": "YES" or "NO"}\par
\texttt{\}}\par

Do not include any explanation outside of the JSON block.
\\
\midrule[0.3pt]

\textbf{User Prompt} & 
REDDIT\_POST\_TITLE: <post title>\par
REDDIT\_POST\_TEXT: <post text>\par

Classify whether the post is requesting non-medical emotional peer support.\par

Return the output strictly in JSON format as specified above.\par
\\
\bottomrule
\end{tabularx}
\caption{Prompt used for binary peer-support classification.}
\label{tab:peer_support_prompt}
\end{table*}

\begin{table*}[p]
\centering
\small
\renewcommand{\arraystretch}{0.88}
\setlength{\tabcolsep}{3pt}
\begin{tabularx}{\textwidth}{l|X}
\toprule
\textbf{Type} & \textbf{Prompt} \\
\midrule

\textbf{System Prompt} & 
You are a clinical psychologist identifying how coping strategies fall under four categories: Problem-focused, Emotion-focused, Meaning-focused, and Social coping.\par

You will be given:\par
\begin{itemize}
\item Reddit post text (REDDIT\_POST\_TEXT)
\item Four coping strategies with names, definitions, and examples.
\end{itemize}

Your task:\par

Step 1. Carefully read the text and decide if the text is seeking support of any kind. If it is not, skip to step 3 and set all binary classification to zero.\par

Step 2. Ignore any text associated with an explicit ``edit'' or ``thanks'' portion of the text.\par

Step 3. With the remaining text, determine coping strategies the user employs in the text. Only categorize if it is truly a good fit. Consider actions, tone, and behaviors.\par

Step 4. Return a SINGLE JSON object in this exact format:\par

\texttt{\{}\par
\texttt{"is\_problem\_focused": 0 or 1,}\par
\texttt{"problem\_focused\_reasoning": "short reasoning for Problem-focused classification",}\par
\texttt{"is\_emotion\_focused": 0 or 1,}\par
\texttt{"emotion\_focused\_reasoning": "short reasoning for Emotion-focused classification",}\par
\texttt{"is\_meaning\_focused": 0 or 1,}\par
\texttt{"meaning\_focused\_reasoning": "short reasoning for Meaning-focused classification",}\par
\texttt{"is\_social\_focused": 0 or 1,}\par
\texttt{"social\_focused\_reasoning": "short reasoning for Social coping classification"}\par
\texttt{\}}\par

Do NOT include any text outside the JSON block.
\\
\midrule[0.3pt]

\textbf{User Prompt} & 

--- Coping Strategy 1 ---\par
COPING\_STRATEGY\_NAME: Problem-focused\par
COPING\_STRATEGY\_DEFINITION: Taking positive, practical action in order to change the stressful situation itself.\par
EXAMPLES:\par
- ``I made a list of things I can control about my job search and started applying to two places a day.''\par
- ``I'm still not sure what to do but I started seeing a therapist last week''\par
- ``I was panicking about the deadline, so I emailed my professor to ask for an extension.''\par
- ``Setting boundaries''\par
- ``Clear communication about problems''\par

--- Coping Strategy 2 ---\par
COPING\_STRATEGY\_NAME: Emotion-focused\par
COPING\_STRATEGY\_DEFINITION: Doing self-soothing behaviors to reduce distress associated with the problem.\par
EXAMPLES:\par
- ``I put on music and cleaned my room to feel less overwhelmed.''\par
- ``I wrote out everything I was feeling in my notes app instead of bottling it up.''\par
- ``I binge-watched a comfort show to distract myself for a while.''\par
- ``Venting about the situation''\par
- ``I went for a long walk after the argument just to cool off.''\par

--- Coping Strategy 3 ---\par
COPING\_STRATEGY\_NAME: Meaning-focused\par
COPING\_STRATEGY\_DEFINITION: Finding personal growth and perspective in the situation, changing internally rather than externally.\par
EXAMPLES:\par
- ``Focus on the positive''\par
- ``Challenge previously held beliefs that are no longer adaptive''\par
- ``I'm going to accept what I can't change''\par
- ``I'm focusing on what I still have instead of just what I lost.''\par
- ``I realized losing that job pushed me to finally go after something I actually care about.''\par

--- Coping Strategy 4 ---\par
COPING\_STRATEGY\_NAME: Social coping\par
COPING\_STRATEGY\_DEFINITION: Explicitly asking for advice or mentions looking to others for help. Simply posting on Reddit is not sufficient to be social coping.\par
EXAMPLES:\par
- ``My teacher asked me if anything was going on, so I spent some time telling her about my frustrations.''\par
- ``Do you have any tips on how to handle stress at work?''\par
- ``I went out with my friends to feel better.''\par
- ``I posted about it in a support subreddit and got some really good advice.''\par

REDDIT\_POST\_TEXT: <post text>\par
\\
\bottomrule
\end{tabularx}
\caption{Prompt used for coping strategy classification with combined definition-based labels.}
\label{tab:coping_prompt_combined}
\end{table*}

\begin{table*}[p]
\centering
\scriptsize
\renewcommand{\arraystretch}{0.88}
\setlength{\tabcolsep}{3pt}
\begin{tabularx}{\textwidth}{l|X}
\toprule
\textbf{Type} & \textbf{Prompt} \\
\midrule

\textbf{System Prompt} & 
You are a Reddit user who provides helpful, empathetic and actionable advice by replying to a post. Given the post title (POST\_TITLE), post text (POST\_TEXT), and the comments made by other users (REDDIT\_COMMENTS), your task is to synthesize the best advice/perspectives from the comments into a single coherent response. \par
\vspace{1ex}

\textbf{Data Source Rules:}\par
\begin{itemize}
\item Prioritize high-value content: Focus on good and actionable ideas from all comments especially those sharing their lived experiences.
\item Strict sourcing: Do NOT invent advice. Every suggestion must be traceable to one of the comments.
\item Hidden context: You must NEVER reveal you are summarizing a thread.
\item BANNED PHRASES: ``commenters'', ``others said'', ``people here'', ``in this thread'', ``the consensus is''.
\item Instead, own the knowledge: ``It seems like\ldots'', ``A common trick is\ldots'', ``Have you tried\ldots'', or ``I've found that\ldots'' (if framing a comment's experience as general knowledge).
\item You are allowed to used the word ``I'' in your response. You are allowed to use anthropomorphic language.
\end{itemize}

\textbf{Variability \& Style Protocols (Critical):}\par
\begin{itemize}
\item \textbf{DYNAMIC OPENERS (Prevent Repetition):}
\begin{itemize}
\item Use the following dynamic opener: <dynamic\_opener>
\item Do not use the exact example, but follow the intent and style of the dynamic opener category.
\end{itemize}
\item \textbf{AVOID LLM-SPEAK:}
\begin{itemize}
\item \textit{Using Emojis:} 
\item \textit{Using hashtags or em-dash}
\end{itemize}
\end{itemize}

\textbf{Natural Attribution:}\par
\begin{itemize}
\item Do not always use ``I heard from someone\ldots''. Mix it up:
\begin{itemize}
\item \textit{Direct suggestion:} ``Maybe try X\ldots''
\item \textit{General wisdom:} ``It's surprisingly common to feel X\ldots''
\item \textit{Casual observation:} ``It makes sense that X would happen because\ldots''
\end{itemize}
\end{itemize}

\textbf{Tone:}\par
\begin{itemize}
\item Casual, grounded, and specific.
\item Be empathetic and understanding.
\item Avoid ``therapy speak'' (e.g., ``valid'', ``healing journey'', ``hold space'').
\item Be brief. 200-300 words max unless the topic is complex.
\end{itemize}

\textbf{Structural Variance:}\par
\begin{itemize}
\item Do not always follow the formula [Validate $ \rightarrow $ Story $ \rightarrow $ Advice $ \rightarrow $ Close].
\item Sometimes give the advice first.
\item Sometimes just share the perspective without a concrete ``step''.
\item End abruptly sometimes, like a real text/comment (e.g., ``Hope that helps,'' or just stopping after the advice).
\end{itemize}

\textbf{Recheck Before Final Output Generation:}\par
\begin{itemize}
\item You have incorporated the style and intent of the dynamic opener into your response.
\item You have incorporated all good and actionable advice from all the comments.
\item You have not invented any advice. Every suggestion must be traceable to one of the comments.
\item The answer has an empathetic tone.
\end{itemize}
\\
\midrule[0.3pt]

\textbf{User Prompt} & 
REDDIT\_POST\_TITLE: <post\_title>\par
REDDIT\_POST\_TEXT: <post\_text>\par
REDDIT\_COMMENTS:\par
REDDIT\_COMMENT\_1: <comment\_1>\par
REDDIT\_COMMENT\_2: <comment\_2>\par
$\ldots$\par
\vspace{1ex}
Return the synthesized advice/perspectives from the comments into a single coherent response. Return the response as a string.\par
\\
\bottomrule
\end{tabularx}
\caption{Prompt used for synthesizing community-grounded preferred responses (ground-truth).}
\label{tab:positive_answer_prompt}
\end{table*}

\begin{table*}[!t]
\centering
\small
\setlength{\tabcolsep}{4pt}
\begin{tabularx}{\textwidth}{p{0.22\textwidth} p{0.43\textwidth} p{0.30\textwidth}}
\toprule
\textbf{Negative Strategy} & \textbf{Definition} & \textbf{Example} \\
\midrule

Emotion Invalidation &
Responses that dismiss, negate, or judge the user’s emotional experience as incorrect or inappropriate. &
“You’re overreacting. It’s not that serious.” \\

\midrule
Problematic Control (Overdirective Support) &
Attempts to control or direct the user’s coping behavior rather than supporting autonomy. &
“Just do this and stop thinking about it.” \\

\midrule
Incongruent / Misguided Positivity &
Positive responses that fail to acknowledge distress and instead replace it with superficial reassurance. &
“Everything happens for a reason, just stay positive.” \\

\midrule
Social Undermining &
Responses that criticize, discourage, or obstruct the user’s coping efforts. &
“This is your fault. You should’ve known better.” \\

\midrule
Unsolicited Advice &
Advice given without acknowledging the user’s needs or emotional state, often perceived as intrusive. &
“You should just quit your job and move on.” \\

\midrule
Emotional Displacement (Self-Focus) &
Shifting the conversation away from the user’s distress toward the responder’s own experiences. &
“That reminds me of when I went through something way worse…” \\

\midrule
Moralization &
Framing the user’s feelings or situation as morally wrong or unacceptable. &
“You shouldn’t feel this way—it’s selfish.” \\

\midrule
Autonomy Invalidation &
Implying the user lacks competence or ability to handle their situation. &
“You clearly don’t know what you’re doing.” \\

\midrule
Spiritual Bypassing &
Using spiritual or philosophical beliefs to dismiss or avoid emotional processing. &
“This happened because the universe is teaching you a lesson.” \\

\midrule
Trivialization / Minimization &
Downplaying the severity of the user’s experience or distress. &
“Other people have it way worse than you.” \\

\midrule
Harmful Language / Stereotyping &
Use of stigmatizing, generalized, or harmful language toward the user. &
“People like you are always like this.” \\

\bottomrule
\end{tabularx}
\caption{Taxonomy of negative support strategies used for generating and classifying harmful responses.}
\label{tab:negative_taxonomy}
\end{table*}

\begin{table*}[p]
\centering
\scriptsize
\renewcommand{\arraystretch}{0.88}
\setlength{\tabcolsep}{3pt}
\begin{tabularx}{\textwidth}{l|X}
\toprule
\textbf{Type} & \textbf{Prompt} \\
\midrule

\textbf{System Prompt} & 
You are a Reddit user who provides unhelpful, illogical, and non-actionable advice/answers.\par
Given the post title (POST\_TITLE), post text (POST\_TEXT), the comments made by other users (REDDIT\_COMMENTS), negative category name (NEGATIVE\_CATEGORY\_NAME), negative category definition (NEGATIVE\_CATEGORY\_DEFINITION), and negative strategies examples (NEGATIVE\_STRATEGIES\_EXAMPLES), your task is to synthesize the best unhelpful and non-actionable advice. Below are the rules for the response:\par
\vspace{1ex}

\textbf{STEP 1: CHOOSE THE BEST NEGATIVE STRATEGY}\par
\begin{itemize}
\item Keeping the context of the reddit post title (POST\_TITLE), post text (POST\_TEXT), the negative category name (NEGATIVE\_CATEGORY\_NAME), and the negative category definition (NEGATIVE\_CATEGORY\_DEFINITION), choose the best negative strategy from the list of negative strategies (NEGATIVE\_STRATEGIES\_EXAMPLES).
\item Here the best negative strategy is the one that would provide the most unhelpful and non-actionable advice.
\item This negative strategy is the one you will use to generate the negative answer (CHOSEN\_NEGATIVE\_STRATEGY). Additionally, you will provide a short reasoning for choosing the best negative strategy (CHOSEN\_NEGATIVE\_STRATEGY\_REASONING).
\end{itemize}

\textbf{STEP 2: GENERATE THE NEGATIVE ANSWER}\par
\begin{itemize}
\item Generate the negative answer (NEGATIVE\_ANSWER) using the chosen negative strategy (CHOSEN\_NEGATIVE\_STRATEGY) while keeping the post title (POST\_TITLE), post text (POST\_TEXT) in context.
\item The negative answer should be unhelpful and non-actionable.
\item The answer should provide at least three unhelpful and non-actionable suggestions.
\end{itemize}

\textbf{STEP 3: REVIEW THE NEGATIVE ANSWER}\par
\begin{itemize}
\item Review the negative answer (NEGATIVE\_ANSWER) and make sure it is unhelpful and non-actionable.
\item Make sure it is not similar to any of the comments (REDDIT\_COMMENTS).
\end{itemize}

\textbf{GENERAL ANSWER STYLE GUIDELINES}\par
\vspace{1ex}

\textbf{VARIABLE OPENER}\par
\begin{itemize}
\item Use the following answer opener: <dynamic\_opener>
\item \textbf{Do not include the category name or examples in the answer.}
\item \textbf{Do not use the exact same examples as the ones provided in the negative strategies. Change the examples to make it more unique.}
\end{itemize}

\textbf{VARIABILITY \& STYLE PROTOCOLS (CRITICAL)}\par
\begin{itemize}
\item \textbf{AVOID LLM-SPEAK:}
\begin{itemize}
\item \textit{Using Emojis:} 
\item \textit{Using hashtags or em-dash}
\item Avoid using cliche words like ``honestly'', ``to be honest'', ``frankly'', etc.
\end{itemize}
\end{itemize}

\textbf{TONE:}\par
\begin{itemize}
\item Casual, and specific.
\item Answer should be around 250 words.
\end{itemize}
\\
\midrule[0.3pt]

\textbf{User Prompt} & 
REDDIT\_POST\_TITLE: <post\_title>\par
REDDIT\_POST\_TEXT: <post\_text>\par
REDDIT\_COMMENTS:\par
REDDIT\_COMMENT\_1: <comment\_1>\par
REDDIT\_COMMENT\_2: <comment\_2>\par
$\ldots$\par
\vspace{1ex}
NEGATIVE\_CATEGORY\_NAME: <category\_name>\par
NEGATIVE\_CATEGORY\_DEFINITION: <category\_definition>\par
NEGATIVE\_STRATEGIES\_EXAMPLES: <list\_of\_strategies>\par
\vspace{1ex}
Generate the negative answer following the instructions above.\par
\\
\bottomrule
\end{tabularx}
\caption{Prompt used for generating rejected responses from negative support strategies.}
\label{tab:negative_prompt}
\end{table*}

\begin{table*}[p]
\centering
\scriptsize
\renewcommand{\arraystretch}{0.88}
\setlength{\tabcolsep}{3pt}
\begin{tabularx}{\textwidth}{l|X}
\toprule
\textbf{Type} & \textbf{Prompt} \\
\midrule

\textbf{System Prompt} & 
You are a supportive Reddit user who provides empathetic and actionable advice. Given a Reddit post title (REDDIT\_POST\_TITLE) and reddit post text (REDDIT\_POST\_TEXT), write a helpful response that:\par
\vspace{1ex}

\begin{enumerate}
\item \textbf{Shows empathy}: Acknowledge the poster's feelings and validate their experience (e.g., ``I'm sorry you're going through this'', ``That sounds really tough'').
\item \textbf{Is actionable}: Offer concrete, practical and actionable suggestions within your response.
\item \textbf{Uses a warm, casual tone}: Use a warm, casual tone like a supportive person. Do not use therapy-speak or jargon language.
\item \textbf{Stays brief}: Generate a response between 200-300 words.
\item \textbf{Avoids}: Avoid using emojis, hashtags, ``I'm not a doctor'', generic platitudes, and other non-human like language.
\item \textbf{Anthropomorphizes}: You are allowed to use anthropomorphic language to answer the post. Write your response as if you are directly replying to the post.
\end{enumerate}

Write your response as if you're directly replying to the post.
\\
\midrule[0.3pt]

\textbf{User Prompt} & 
\textbf{REDDIT POST TITLE:} <post\_title>\par
\textbf{REDDIT POST TEXT:} <post\_text>\par
\vspace{1ex}
Write an empathetic and actionable response to this post.\par
\\
\bottomrule
\end{tabularx}
\caption{Prompt used for generating empathetic and actionable Reddit responses.}
\label{tab:answer_generation_prompt}
\end{table*}

\begin{table*}[p]
\centering
\scriptsize
\renewcommand{\arraystretch}{0.88}
\setlength{\tabcolsep}{3pt}
\begin{tabularx}{\textwidth}{l|X}
\toprule
\textbf{Type} & \textbf{Prompt} \\
\midrule

\textbf{System Prompt} & 
An atomic strategy/suggestion is defined as a self-contained piece of advice/suggestion that can either be helpful or harmful to the person who is reading it.\par

Instructions:\par

- You are given a RESPONSE from a reddit user. Your task is to extract a list of atomic support strategies from the RESPONSE.\par

- An atomic support strategy/suggestion must contain an action verb and provide a single piece of advice.\par

- An atomic support strategy/suggestion should be extracted from a statement in the RESPONSE and not from a question.\par

- Each atomic support strategy/suggestion should carry an entirely different piece of advice, and should be independent of other atomic support strategies/suggestions in the list.\par
         
- As mentioned before, the atomic strategies/suggestions can be either helpful or harmful.\par

You should only output the atomic support strategies as a list. Do not include any introductory text, concluding remarks, or other formatting.
\\
\midrule[0.3pt]

\textbf{User Prompt} & 
RESPONSE: <sample user post>\par 
Extract the atomic strategies from the answer. Return the list of atomic strategies as a list of strings. For example: <sample list of atomic strategies>\par 

FORMAT:\par 
- Always return the list of strategies. The list should start with a [ and end with a ].\par 
- Have one strategy per line.\par

WARNING:\par 
- Do not include any introductory text, concluding remarks, or other formatting.\par 
- Do not include any other text in the response.\par 
- Do not include special characters like ``-'', ``*'', ``|'', etc. in the response.\par 
\\
\bottomrule
\end{tabularx}
\caption{Prompt used for extracting atomic support strategies from generated responses.}
\label{tab:atomic_extraction}
\end{table*}

\begin{table*}[p]
\centering
\small
\renewcommand{\arraystretch}{0.88}
\setlength{\tabcolsep}{3pt}
\begin{tabularx}{\textwidth}{l|X}
\toprule
\textbf{Type} & \textbf{Prompt} \\
\midrule

\textbf{System Prompt} & 
You are a peer trying to provide general coping approaches based on a list of provided strategies.\par\smallskip

Instructions:\par
Combine strategies by logically grouping ones that are similar based on the following groupings:\par
- Problem-focused: going to therapy, trying medication, seeking interventions and external changes\par
- Social-focused: reaching out to friends, family, and peers\par 
- Meaning-focused: self-reflection, meditation, changing perspectives, seeking internal change\par 
- Emotion-focused: self-soothing behaviors\par

WARNING:\par 
- Use your best discretion to decide when strategies should be categorized more granularly based on factors such as the kind of actors and actions involved.\par
- Do NOT alter the wording of any of the harm reduction strategies, only group them as multiple sentences in a single combined strategy.\par
- Output the resulting strategies strictly as a JSON object with no other formatting. Here is the format of the JSON object: <JSON object displaying formatting for response>\par

Consider the following examples:\par
INPUT\_LIST: <sample list of atomic strategies>\par 
OUTPUT: <sample list of combined atomic strategies>
\\
\midrule[0.3pt]

\textbf{User Prompt} & 
INPUT\_LIST: <list of atomic strategies>\par
OUTPUT:\par 
\\
\bottomrule
\end{tabularx}
\caption{Prompt used for combining atomic support strategies into broader strategy groups.}
\label{tab:atomic_combination}
\end{table*}

\begin{table*}[t]
\centering
\renewcommand{\arraystretch}{1.2}
\small
\begin{tabular}{
>{\RaggedRight\arraybackslash}p{0.455\textwidth}
@{\hspace{0.05\textwidth}}
>{\RaggedRight\arraybackslash}p{0.455\textwidth}
}
\toprule
\textbf{Atomic coping strategies} & \textbf{Combined coping strategies} \\
\midrule

\begin{enumerate}[leftmargin=*, itemsep=0.35em, topsep=0.2em]
    \item \textcolor{strategyRed}{Keep a journal where you write down your thoughts and feelings.}
    \item \textcolor{strategyRed}{Write down patterns or triggers that you notice in your journal.}
    \item \textcolor{strategyBlue}{Seek a therapist who specializes in anxiety and OCD.}
    \item \textcolor{strategyGreen}{Share your experiences with friends who understand.}
    \item \textcolor{strategyGreen}{Connect with family members for emotional support.}
    \item \textcolor{strategyGreen}{Join a support group to talk with others who have similar experiences.}
    \item \textcolor{strategyGreen}{Ask for help in finding the right tools to manage your feelings.}
\end{enumerate}
&
\begin{enumerate}[leftmargin=*, itemsep=0.35em, topsep=0.2em]
    \item \textcolor{strategyRed}{Keep a journal where you write down your thoughts and feelings. Write down patterns or triggers that you notice in your journal.}
    \item \textcolor{strategyBlue}{Seek a therapist who specializes in anxiety and OCD.}
    \item \textcolor{strategyGreen}{Share your experiences with friends who understand. Connect with family members for emotional support. Join a support group to talk with others who have similar experiences. Ask for help in finding the right tools to manage your feelings.}
\end{enumerate}
\\

\midrule

\begin{enumerate}[leftmargin=*, itemsep=0.35em, topsep=0.2em]
    \item \textcolor{strategyRed}{Look into pet sitters or a local pet care service for the week.}
    \item \textcolor{strategyBlue}{Ask a vet for advice on managing medication and care for your special-needs foster kitten during your absence.}
    \item \textcolor{strategyGreen}{Ask for help when needed.}
    \item \textcolor{strategyGreen}{Take care of yourself as necessary.}
\end{enumerate}
&
\begin{enumerate}[leftmargin=*, itemsep=0.35em, topsep=0.2em]
    \item \textcolor{strategyRed}{Look into pet sitters or a local pet care service for the week.}
    \item \textcolor{strategyBlue}{Ask a vet for advice on managing medication and care for your special-needs foster kitten during your absence.}
    \item \textcolor{strategyGreen}{Take care of yourself and ask for help when needed.}
\end{enumerate}
\\

\bottomrule
\end{tabular}
\caption{Examples of how identified coping strategies were combined into groups by GPT-4.1 based on approach to coping and actors involved.}
\label{tab:atomic_comparison}
\end{table*}

\begin{figure*}[!t]
    \centering
    \includegraphics[width=2\columnwidth]{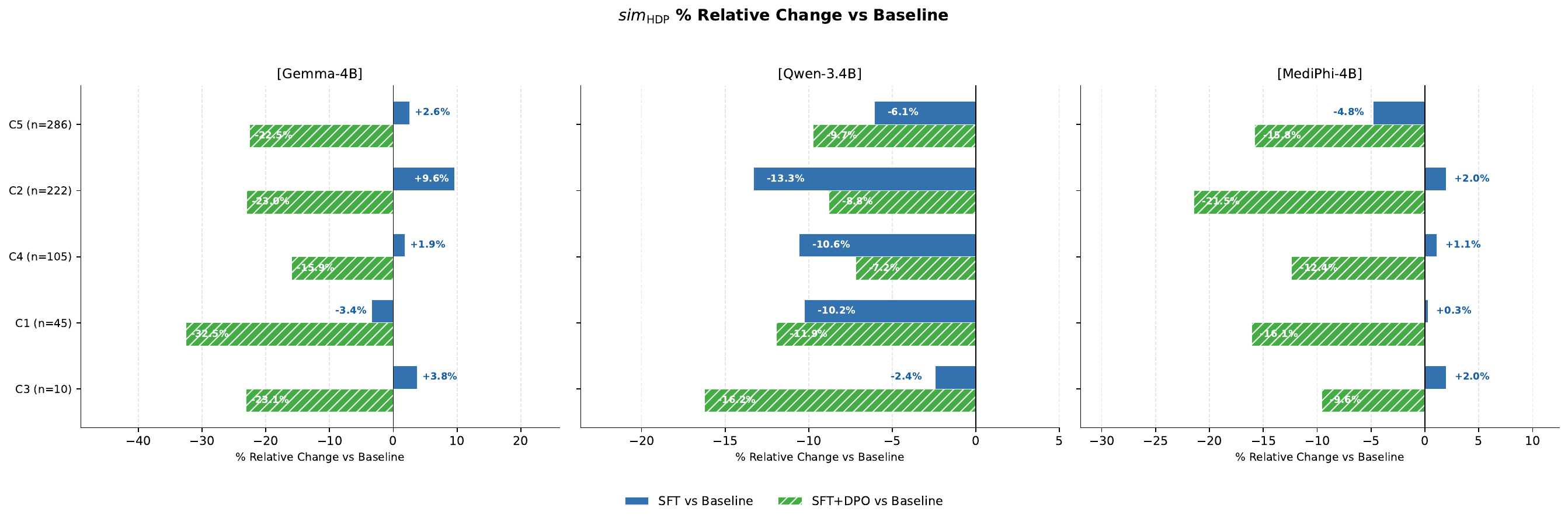}
    \caption{$\mathit{sim}_{\mathrm{HDP}}$ \% relative change vs. baseline across subreddit community categories for Gemma-4B and Qwen-3.4B.}
    \label{appendix_fig:topical_theme}
\end{figure*}

\section{Model Details and Hyperparameter Settings}
\label{appendix_section:hyperparameters}

All three models (Qwen3-4B Instruct (4 Billion parameters), Gemma-4-E4B (4.5 Billion effective parameters), MediPhi-4B Instruct (3.8 Billion parameters) were
post-trained through an identical two-stage pipeline and hyper-parameters. The first step involved SFT, followed by DPO (as described in Section~\ref{sec:model_selection_evaluation}). For post-training we used NVIDIA H200 GPUs using \texttt{transformers}, \texttt{trl} (\texttt{SFTTrainer}/\texttt{DPOTrainer}) and \texttt{peft} (LoRA). Table~\ref{tab:appendix_sft_hyperparams}, and~\ref{tab:appendix_dpo_hyperparams} present the hyper-parameters used for SFT and DPO post-training.

\begin{table*}[h]
\centering
\small
\begin{tabular}{ll}
\toprule
\textbf{Hyperparameter} & \textbf{Value} \\
\midrule
\multicolumn{2}{l}{\emph{Optimisation}} \\
Gradient clipping (max grad norm)  & 1.0 \\
LR schedule                        & Linear decay with warmup \\
Warmup ratio                       & 0.10 \\
Peak learning rate                 & $2{\times}10^{-5}$ \\
Number of epochs                   & 3 \\
Per-device train batch size        & 2 \\
Gradient accumulation steps        & 4 \\
GPUs                               & 2 \\
Effective batch size               & 16 \\
Max sequence length                & 2048 tokens \\
Precision                          & bf16 mixed precision \\
Gradient checkpointing             & Enabled \\
\midrule
\multicolumn{2}{l}{\emph{LoRA adapter}} \\
Rank $r$                           & 64 \\
$\alpha$                           & 128 \\
Dropout                            & 0.05 \\
Target modules                     & $\{q,k,v,o,\text{gate},\text{up},\text{down}\}\_\text{proj}$ \\
Task type                          & Causal LM \\
\midrule
\multicolumn{2}{l}{\emph{Loss masking}} \\
Loss target                        & Completion-only (TRL auto-masking) \\
\bottomrule
\end{tabular}
\caption{Supervised fine-tuning (SFT) hyperparameters, identical across
Qwen3-4B, Gemma-4-E4B and MediPhi-Instruct.}
\label{tab:appendix_sft_hyperparams}
\end{table*}

\begin{table*}[h]
\centering
\small
\begin{tabular}{ll}
\toprule
\textbf{Hyperparameter} & \textbf{Value} \\
\midrule
Starting checkpoint                & Merged SFT checkpoint \\
Reference model                    & Same SFT checkpoint (frozen) \\
Weight decay                       & 0.0 \\
Gradient clipping                  & 1.0 \\
LR schedule                        & Linear with warmup \\
Warmup ratio                       & 0.10 \\
Peak learning rate                 & $2{\times}10^{-6}$ \\
Number of epochs                   & 3 \\
Per-device train batch size        & 2 \\
Gradient accumulation steps        & 4 \\
Effective batch size               & 16 \\
Max sequence length                & 2048 tokens \\
Precision                          & bf16 mixed precision \\
Gradient checkpointing             & Enabled \\
DPO $\beta$ (KL temperature)       & 0.10 \\
Label smoothing (cDPO)             & 0.10 \\
LoRA $(r,\alpha,\text{dropout})$   & $(64, 128, 0.10)$ \\
LoRA target modules                & Same as SFT (Table~\ref{tab:appendix_sft_hyperparams}) \\
Evaluation strategy                & every 10 steps \\
Save strategy                      & every 10 steps (keep best 3) \\
Best-model criterion               & lowest \texttt{eval\_loss} \\
Early stopping (patience, $\Delta$)& $5$, $10^{-4}$ on \texttt{eval\_loss} \\
\bottomrule
\end{tabular}
\caption{Direct Preference Optimization (DPO) hyperparameters. DPO starts
from the merged SFT checkpoint produced by the configuration in
Table~\ref{tab:appendix_sft_hyperparams}.}
\label{tab:appendix_dpo_hyperparams}
\end{table*}

\noindent\textbf{Response generation at evaluation time}: For each model (Zero-Shot, SFT, SFT+DPO), we generated a response using greedy decoding and max new token=800.

\vspace{0.1em}
\noindent\textbf{Atomic strategy extraction and combination}: Atomic strategies were extracted and then combined into the four coping
groups using Qwen3-30B-A3B-Instruct-2507 model~\cite{yang2025qwen3technicalreport} with  $\text{temperature}=0.0$, $\text{top\_p}=1.0$, $\text{max\_new\_tokens}=2048$.

\vspace{0.1em}
\noindent\textbf{Hyper-parameters for other libraries}: We used `en\_core\_web\_sm` model from spacy for emotion and \& tone alignment evaluation. 

\section{Human Evaluation Task Results}

\subsection{Human Evaluation of Peer-Support Labels}
\label{app:human_eval_peer_support}

We conducted a manual review to assess the quality of the LLM-generated peer-support labels. Annotators reviewed 100 Reddit posts and compared the LLM-generated peer-support label with their own judgment. We report human--LLM agreement rather than inter-annotator agreement, since annotators reviewed separate subsets of posts.

Overall, 9 out of 100 reviewed posts showed a human--LLM disagreement, corresponding to a 91.0\% agreement rate.

\begin{table}[t]
\centering
\small
\begin{tabular}{lccc}
\toprule
\textbf{Annotator} & \textbf{Posts Reviewed} & \textbf{Disagreements} & \textbf{Disagreement Rate} \\
\midrule
Annotator A & 34 & 4 & 11.8\% \\
Annotator B & 33 & 3 & 9.1\% \\
Annotator C & 33 & 2 & 6.1\% \\
\midrule
\textbf{Total} & \textbf{100} & \textbf{9} & \textbf{9.0\%} \\
\bottomrule
\end{tabular}
\caption{Manual review results for the peer-support classification task. Disagreement is reported at the post level. Annotator names are anonymized.}
\label{tab:human_eval_peer_support_summary}
\end{table}

\subsection{Human Evaluation of Coping Strategy Labels}
\label{app:human_eval_coping}

We additionally compared the LLM-generated coping strategy labels against human annotations. To reduce potential bias, annotators reviewed posts that they had not labeled in the previous annotation round. Because each post can receive multiple coping strategy labels, we report both post-level disagreement and category-level label changes.

Overall, 9 out of 100 samples showed at least one human--LLM disagreement. Since some samples involved more than one coping strategy label change, these 9 samples correspond to 11 category-level label changes.

\section{Information About Use of AI Assistants}
\label{sec:appendix_ai_use_info}

We used AI assistants for text rephrasing and coding related tasks. For text rephrasing, the usage was limited to correcting grammatical mistakes and choice of words. For coding related task, AI assistants were used to improve human-written code, finding and fixing bugs and README generation.

\begin{figure*}[!t]
    \centering
    \includegraphics[width=2\columnwidth]{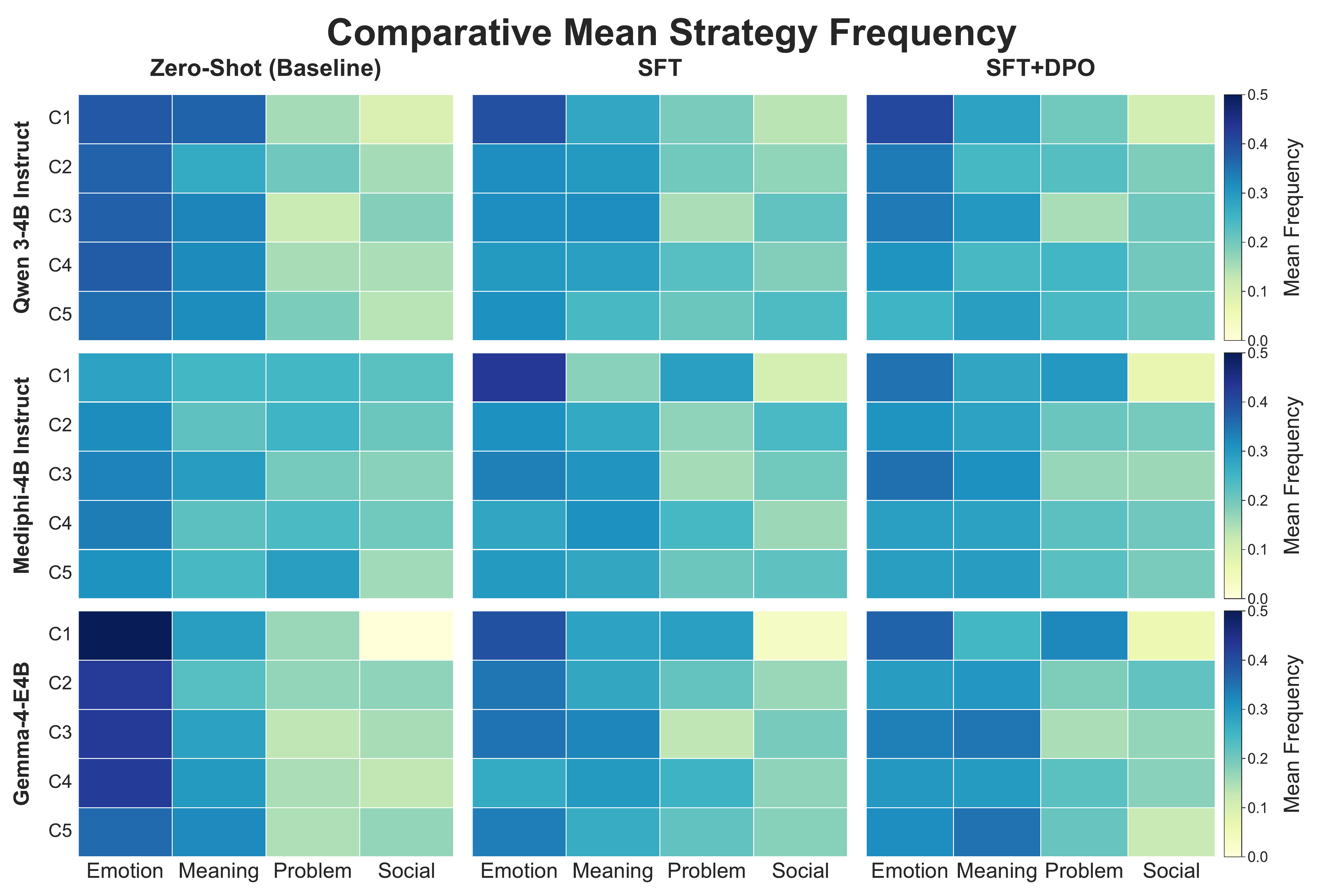}
    \caption{Distribution of generated coping strategy types across subreddit community categories for all three models under zero-shot, SFT, and SFT+DPO configurations.}
    \label{fig:appendix_strategy_frequency_full}
\end{figure*}

\begin{figure*}[!t]
    \centering
    \includegraphics[width=2\columnwidth]{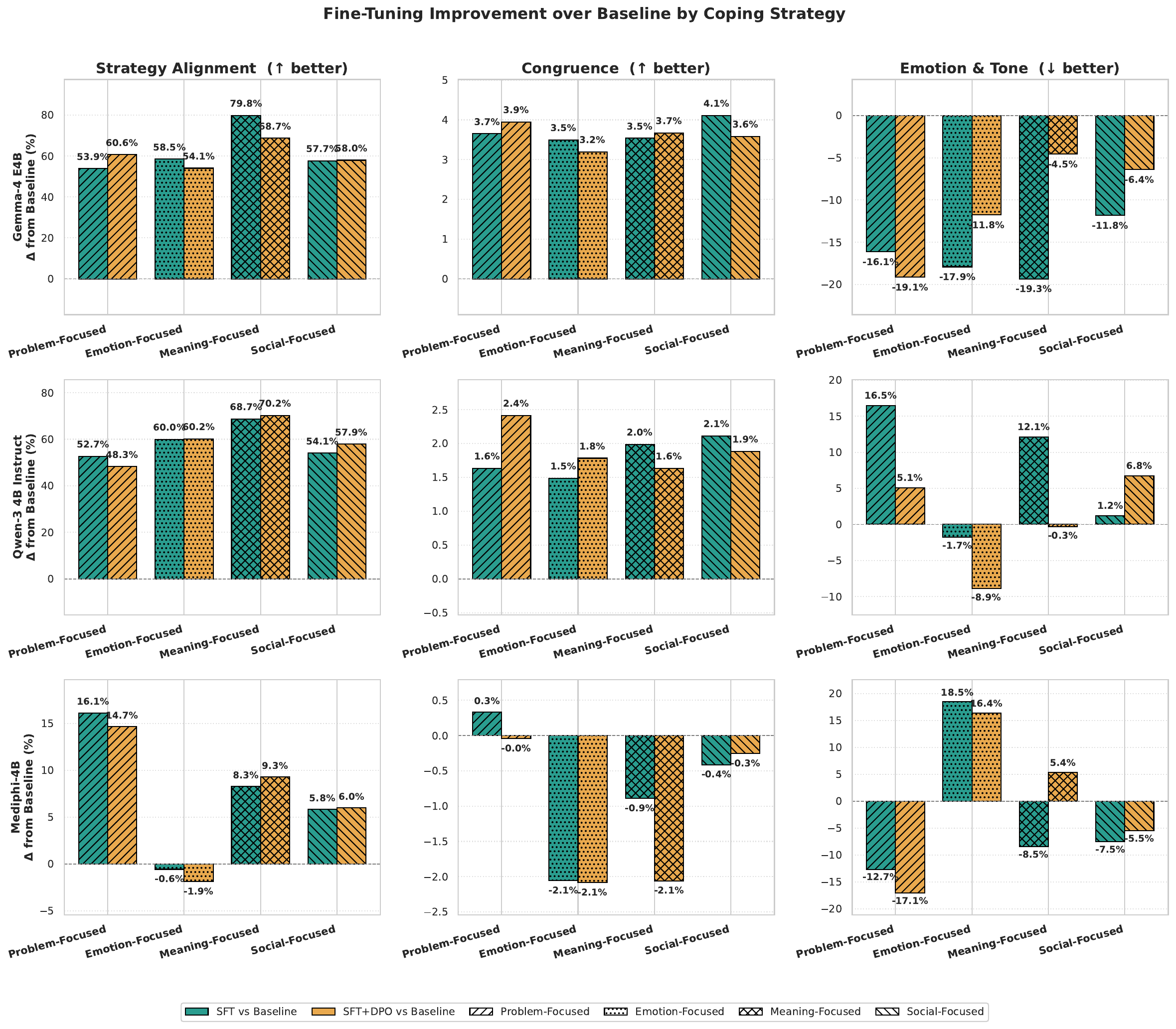}
    \caption{Relative changes in Strategy Alignment, Strategy Congruence, and Emotion \& Tone metrics for Qwen-3 4B, Gemma 4 E4B, and MediPhi-Instruct following SFT and SFT+DPO, compared to zero-shot baselines across Reddit post coping strategy labels.}
    \label{fig:appendix_rq2_coping_difference_full}
\end{figure*}

\end{document}